# Chapter 16

# Memory-Driven Metaheuristics: Improving Optimization Performance

Salar Farahmand-Tabar
Department of Civil Engineering Eng., Faculty of Engineering, University of Zanjan, Zanjan, Iran,

**Abstract.** Metaheuristics are stochastic optimization algorithms that mimic natural processes to find optimal solutions to complex problems. The success of metaheuristics largely depends on the ability to effectively explore and exploit the search space. Memory mechanisms have been introduced in several popular metaheuristic algorithms to enhance their performance. This chapter explores the significance of memory in metaheuristic algorithms and provides insights from well-known algorithms. The chapter begins by introducing the concept of memory, and its role in metaheuristic algorithms. The key factors influencing the effectiveness of memory mechanisms are discussed, such as the size of the memory, the information stored in memory, and the rate of information decay. A comprehensive analysis of how memory mechanisms are incorporated into popular metaheuristic algorithms is presented, and concludes by highlighting the importance of memory in metaheuristic performance and providing future research directions for improving memory mechanisms. The key takeaways are that memory mechanisms can significantly enhance the performance of metaheuristics by enabling them to explore and exploit the search space effectively and efficiently, and that the choice of memory mechanism should be tailored to the problem domain and the characteristics of the search space.

## 1. Introduction

Optimization is a fundamental tool used in various fields to find the best solution to a problem given a set of constraints. An optimization problem involves minimizing or maximizing an objective function subject to constraints. For example, in a manufacturing process, optimization can be used to minimize the production cost or maximize the production output while satisfying the production constraints. Optimization problems can be complex and difficult to solve, especially when the search space is large or there are many constraints. In such cases, traditional optimization methods such as mathematical programming may not be able to find the optimal solution within a reasonable time. Metaheuristics are a class of algorithms that are designed to overcome these challenges.

Metaheuristics are general problem-solving techniques that can be applied to a wide range of optimization problems. Unlike mathematical programming, which requires a specific model and constraints, metaheuristics do not rely on a specific problem structure. Instead, they use iterative search procedures to explore the solution space and find the best solution. The search process of metaheuristics involves generating a set of candidate solutions, evaluating the solutions based on the objective function, and then modifying the solutions to generate a new set of candidates. This process is repeated until the algorithm converges to the best solution or a stopping criterion is met.

There are several types of metaheuristics, each with its own unique approach and characteristics. For example, evolutionary metaheuristics are based on the principles of natural selection such as Genetic Algorithm [1]. Swarm Intelligence are inspired by the social behavior of swarming insects and animals such as Particle swarm optimization [2]. Physic-based metaheuristics are another type of metaheuristics, which are based on physical principles such as energy, forces, and vibrations such as Thermal Exchange [3]. Human-inspired metaheuristics are inspired by human behavior and problem-solving approaches such as Neural Network [4].

Metaheuristics are powerful optimization algorithms that are used to solve complex problems, but they can sometimes get stuck in local optima or fail to converge to the global optima. To overcome these limitations, several techniques have been developed to enhance the performance of metaheuristics. One of the most commonly used techniques is parameter tuning, which involves optimizing the values of the metaheuristic's parameters to improve its performance. Another technique is hybridization, which involves combining two or more metaheuristics to create a more powerful algorithm. This technique can combine the strengths of different metaheuristics to overcome their weaknesses and improve their overall performance.

There are numerous improving features to enhance the performance of the optimization algorithms such as levy flight or chaotic maps [4]. Another feature used in enhanced metaheuristics is memory-based improvement. Memory-based improvement involves storing and reusing the information of previous solutions to improve the performance of the metaheuristic algorithm. This approach is commonly used in metaheuristics that rely on a population-based search, such as genetic algorithms and particle swarm optimization. By storing and reusing the information of previous solutions, memory-based improvement can help the algorithm avoid local optima and improve convergence speed. One example of memory-based improvement is the use of adaptive memory, which involves storing the best solutions found so far and adjusting the search process based on the historical information.

As evident from the literature review, many basic optimization algorithms have several limitations that hinder their ability to solve complex problems efficiently. Some of the common issues include slow convergence speed and the tendency to get trapped in local optima. To overcome these issues, researchers have proposed various mechanisms to enhance the performance of these algorithms. In this chapter, we propose the use of memory-assisted optimization algorithms as an improvement feature. The memory-assisted version of well-known optimization algorithms such as Multi-Verse Optimizer (MVO), Vibrating Particle Search (VPS), Thermal Exchange Optimization (TEO), and Ray optimization (RO) is implemented as an optimization method, with a separate memory component for storing and exchanging the best solutions found so far. The efficiency of utilizing memory in these algorithms is investigated through various benchmark engineering examples, and its effectiveness is compared with that of memory-less versions of the algorithms.

## 2. Background Studies on Memory-Enhanced Metaheuristics

Memory-enhanced metaheuristics have gained increasing attention in recent years due to their ability to improve the performance of optimization algorithms. These methods are designed to incorporate memory mechanisms into conventional metaheuristics, such as Differential Evolution (DE), Genetic Algorithms (GA), Particle Swarm Optimization (PSO), Artificial Bee Colony (ABC), Grey Wolf Optimizer (GWO), Ant Colony Optimization (ACO), and Whale Optimization Algorithm (WOA), etc. The use of memory mechanisms can enhance the search process by storing and utilizing information about previously visited solutions, enabling the algorithm to explore the search space more effectively and converge to better solutions faster. In this section, a comprehensive overview of the background studies on memory-enhanced metaheuristics is provided, covering a range of optimization methods and their applications in various fields. Table 1 summarizes the methods and their applications discussed in this section.

**Table 1.** Background studies on memory-assisted metaheuristics

| Method | Application | Ref. | Method | Application | Ref. |
|---|---|---|---|---|---|
| GA | Laminate composites | [5] | SAGA | Multimodal optimization | [6] |
| GA | Power Generation | [7] | GA | Vehicular Communication | [8,9] |
| NSGA II | Dynamic Problems | [10] | EA | Job Shop Scheduling | [11] |
| DE | Global optimization | [12,13] | HDE | Continuous problems | [14] |
| MODE | Multiobjective Optimization | [15,19] | PSO | Dynamic optimization | [20] |
| QPSO | Large scale problems | [21] | PSO | Feature Selection | [22] |
| BBPSO | No-Linear Functions | [23] | IPSO | Training MLP | [24] |
| BPSO | Discrete benchmark func. | [25] | ACO | Numerical Optimization | [26,27] |
| IACO | Traveling Salesman | [28] | ACO | Dynamic vehicle routing | [29] |
| DA | Engineering problems | [30,31] | FFOA | Image Segmentation | [32,33] |
| IFFOA | Time series forecasting | [34] | ABC | Dynamic optimization | [35,36] |
| GWO | Global optimization | [37] | NFGWO | Optimal load balancing | [38] |
| AFSA | Multi-extreme value func. | [39] | SCA | Global optimization | [40] |
| HS, TS | Global optimization | [41,42] | MVO | Structural optimization | [43] |
| WOA | Smooth path planning | [44] | KMTOA | Global optimization | [45] |
| MMGSO | Benchmark functions | [46] | OWMA | Global optimization | [47] |
| CFA | Global optimization | [48,49] | SLH | Large-Scale p-Median Prob. | [50] |
| BBBC | Data clustering | [51] | GD | Global optimization | [52] |
| HMBOA | Dynamic environments | [53] | P-PDA | Image processing | [54] |
| CSA | Global optimization | [55] | TRA | Unconstrained optimization | [56] |
| CA | Combinatorial optimization | [57] | AIS | Dynamic optimization | [58] |

The applications range from global and engineering optimization to machine learning related applications. The incorporation of memory mechanisms into these metaheuristics has been found to improve the performance of the algorithms, allowing for more effective exploration of the search space and faster convergence to better solutions. Overall, the use of memory-enhanced metaheuristics is a promising area of research with the potential to enhance optimization performance across a wide range of applications.

## 3. Memory assignment (multi-elite strategy)

To enhance the performance of an algorithm without introducing additional computational costs, an approach called the multi-elite strategy can be employed, which involves utilizing a separate memory to store historical best solutions and their corresponding fitness values. This strategy differs from the elitist strategy typically utilized by standard algorithms like

GA [59, 60], which rely on a single elite strategy. In the multi-elite strategy, a certain number of the best solutions found so far (based on the designated memory size) are preserved in a dedicated memory. In each iteration, these elite solutions are exchanged with an equal number of the worst solutions. In this study, the memory size, which represents the number of elite solutions stored, is set as 1/5 of the total number of universes. By incorporating this memory into each iteration (Algorithm 1), undesirable and poor-performing solutions can be replaced with the desired elite solutions. The memory is continuously updated with new solutions, and if the fitness of the new solutions surpasses that of the stored elite solutions, they are exchanged, along with their corresponding fitness values. This modification enables the algorithm to avoid suboptimal solutions and achieve faster and more efficient convergence.

**Algorithm 1. Pseudocode of Assigning Memory**

```
Start
% Saving more than one global best and related solutions
for i = 1: NU
  for j = 1: NU/5            % Memory size: 20 percent of search agents
    if Fit(i) < Fit_M(j)
      U_M(j,:) = U(i,:);     % Best solutions of the memory
      Fit_M(j) = Fit(i);     % Best fitness values of the memory
    end
  end
end
% Worst 20% of search agents are changed in the main loop of the algorithm
```

## 4. Memory-Enhanced Metaheuristics

In this section, the metaheuristics utilized in the study are introduced to demonstrate the effectiveness of memory-enhanced optimization. The chosen algorithms are well-known and widely used in the optimization literature, including the Biogeography-Based Optimization (BBO), Krill Herd Algorithm (KHA), and Thermal Exchange Optimization (TEO). These algorithms are categorized under swarm intelligence and physics-based metaheuristics and have been successfully applied in various optimization problems, including engineering, computer science, and other fields. In this study, it is aimed to enhance these algorithms by incorporating a memory mechanism that stores historical best solutions to improve their overall performance in terms of convergence speed, quality of solution, and reliability. These algorithms are applied to optimize the problems to be evaluated their performance with and without memory enhancement.

## 4.1. Biogeography-Based Optimization

BBO primarily employs species migration and mutation models in the field of biogeography to address optimization issues. In BBO [61], individual solutions are referred to as "habitats", and their quality is evaluated using a Habitat Suitability Index (HSI). The Suitability Index Variables (SIVs) represent the factors that define the habitability of a habitat. BBO primarily relies on migration and mutation processes to explore and discover the most optimal solution.

### 4.1.1. Migration Operator

In the BBO (Biogeography-Based Optimization) algorithm, a high HSI (Habitat Suitability Index) indicates a good solution, analogous to a habitat with abundant species. Such habitats exhibit high emigration rates (species leaving the habitat) and low immigration rates (species entering the habitat), and vice versa. The migration operator in the algorithm aims to facilitate the exchange of information among different solutions. In this context, good solutions tend to share their favorable characteristics with poor solutions, while poor solutions are more receptive to adopting beneficial features from good solutions. Each habitat within the algorithm has its specific emigration rate ($\mu$), and immigration rate ($\lambda$) which are computed as follows:

$$\lambda_k = I\left(1 - \frac{N_k}{N}\right), \quad \mu_k = E\left(\frac{N_k}{N}\right) \qquad (1)$$

In these equations, $I$ represents the maximum immigration rate, $E$ denotes the maximum emigration rate, $N_k$ corresponds the number of species of the habitat $H_k$, and $N$ represents the maximum number of species. It's worth noting that while the given equations present a simple linear model for migration, in practice, more complex and nonlinear models are often utilized in the BBO algorithm. The migration operator in the BBO algorithm modifies the SIVs of a habitat by incorporating features from other advantageous habitats. This process can be expressed as follows:

$$H_i(SIV) \leftarrow H_k(SIV) \qquad (2)$$

In the given expression, $H_i$ represents the immigration habitat, while $H_k$ denotes the emigration habitat. The emigration habitat $H_k$ is chosen using the roulette wheel selection method.

### 4.1.2. Mutation Operator

In BBO, sudden events can lead to significant changes in the characteristics of a habitat, resulting in alterations to its HSI and the number of species present. The probability of species number in BBO is directly related to the mutation rate of a habitat. More specifically, the mutation rate ($m_i$) is determined by the probability ($p_i$) of the species number, and it can be mathematically expressed as follows:

$$m_i = m_{max}(1 - \frac{p_i}{p_{max}}) \tag{3}$$

In the provided equation, $m_{max}$ represents the maximum mutation rate, which is a parameter defined by the user. The calculation of $p_i$ follows a specific computation method, and $p_{max}$ corresponds to the maximum value among all $p_i$ probabilities. The mutation process can be performed as follows:

$$H_i(SIV_j) \leftarrow lb_j + rand * (ub_j - lb_j) \tag{4}$$

Here, $H_i$ represents the mutation habitat. For each SIV in $H_i$, denoted by $j$ ranging from 1 to $D$ (where $D$ is the number of decision variables), the mutation operation is conducted. The lower and upper boundary values of the $j$th SIV in $H_i$ are represented by $lb_j$ and $ub_j$, respectively. The mutation process involves modifying the SIV by adding a uniformly distributed random real number, rand, between 0 and 1.

In order to maintain the most optimal solutions throughout the search process, BBO utilizes the strategy of elitism. This approach involves several steps: during each iteration, after executing operations such as migration and mutation, the population is sorted. Following this, a number of the least favorable habitats are replaced with some of the top-performing solutions that were preserved from previous iterations. Once this replacement is completed, the population is sorted once more. To summarize, the BBO algorithm follows the following steps:

*Step 1:* set the parameters and initialize the random population
*Step 2:* compute each habitat and sort the population considering their $HSIs$ in descending order
*Step 3:* evaluate the immigration, emigration, and mutation rates and keep the elitists
*Step 4:* perform the migration and mutation operator by Eq. (2 and 4)
*Step 5:* restrict the boundary of each new solution
*Step 6:* compute HSI of each habitat's and sort the population in descending order

*Step 7:* replace several worst habitats with elitists ones and sort the population in descending order

*Step 8:* if the termination criterion is satisfied, output the optimum solution; otherwise, return to Step 3

## 4.2.   Krill Herd Algorithm

Krill swarms, a marine species studied by humans, exhibit a tendency to form clusters. When these krill swarms encounter natural predators or disturbances, some individuals may be lost or displaced, leading to a reduction in population density. To restore the original state, krill swarms exhibit two main behaviors: increasing population density and searching for food. Inspired by these behaviors, researchers have proposed a novel heuristic algorithm called the Krill Herd Algorithm (KHA). The KHA algorithm aims to solve global optimization problems by simulating the clustering and foraging behaviors observed in krill swarms.

In the Krill Herd Algorithm (KHA) [62], every individual krill represents a potential solution for the optimization problem at hand. The two goals of increasing population density and finding food are considered as the driving forces for the optimization problem. The process of re-aggregating individual krill represents the algorithm's search for the optimal solution. The location of each krill evolves over time, primarily influenced by the following three factors:

- Movement induced by other krill individuals
- Foraging motion
- Random diffusion

In KHA, the Lagrangian model is applied to tackle decision problems that involve multiple dimensions:

$$\frac{dX_i}{dt} = N_i + F_i + D_i \qquad (5)$$

where $N_i$ represents the induced motion of other krill individuals; $F_i$ denotes the Foraging activity and $D_i$ corresponds the physical diffusion.

### 4.2.1. Movement Induced by Other Krill Individuals

To facilitate the collective migration of the population, every krill individual in the KHA algorithm interacts with one another, fostering a high population density. The direction of movement (denoted as $\alpha_i$) for each krill is influenced by three factors: the influence of neighboring individuals (local effect), the impact of the optimal individual (target effect), and the repulsion effect from the population as a whole (repulsive effect). The movement induced by other krill individuals ($N_i$) for a given krill can be expressed as follows:

$$N_i = N^{max}\alpha_i + \omega_n N_i^{old} \tag{6}$$

$$\alpha_i = \alpha_i^{local} + \alpha_i^{target} \tag{7}$$

In the equation, $N^{max}$ represents the maximum induced speed, and $N_i^{old}$ denotes the previously induced motion for the krill individual. The inertia weight of the motion, $\omega_n$, takes a value between 0 and 1 and represents the influence of the krill's previous motion on the current movement. $\alpha_i^{local}$ and $\alpha_i^{target}$ represent the local effect and target effect, respectively. The local effect, induced by neighboring krill individuals, can be interpreted as an attractive or repulsive tendency. It is determined by the following expression:

$$\alpha_i^{local} = \sum_{j=1}^{NN} \hat{K}_{i,j}\hat{X}_{i,j} \tag{8}$$

$$\hat{X}_{i,j} = \frac{X_j - X_i}{\|X_j - X_i\| + \varepsilon}, \qquad \hat{K}_{i,j} = \frac{K_i - K_j}{K^{worst} - K^{best}} \tag{9}$$

where $NN$ is the number of neighbors, $X$ represents the related position, and K represents the fitness value of the krill individual. $K^{worst}$ and $K^{best}$ represent the worst and best fitness values observed among the krill herds thus far. Additionally, $\varepsilon$ is a small positive value introduced to avoid singularities and ensure stability in the calculations. The calculation of the local effect in the KHA algorithm involves determining the neighbors of a krill individual based on their sensing distance. The sensing distance determines which other krill individuals are considered as neighbors. It is defined using the following formula:

$$d_i = \frac{1}{5NP} \sum_{j=1}^{NP} \|X_i - X_j\| \tag{10}$$

where NP denotes the size of the population. The movement of each krill is influenced by the global optimal solution, which serves as the target direction. This influence on movement can be described as follows:

$$\alpha_i^{target} = C^{best}\widehat{K}_{i,best}\widehat{X}_{i,best}, \quad C^{best} = 2\left(rand_1 + \frac{g}{g_{max}}\right) \tag{11}$$

In the equation, $rand_1$ represents a random variable uniformly distributed between 0 and 1. The variables $g$ and $g_{max}$ correspond to the number of current iteration and the maximum iterations, respectively.

### 4.2.2. Foraging Motion

The population's search for food in the foraging process involves estimating the desired resource based on the fitness distribution of the krill individuals. The location of the resource is determined using the concept of the "center of mass" from physics:

$$X^{food} = \frac{\sum_{i=1}^{NP} \frac{1}{K_i} X_i}{\sum_{i=1}^{NP} \frac{1}{K_i}} \tag{12}$$

Two primary factors influence the foraging behavior of krill: the current location of the food source and its previous location. This relationship can be expressed as follows:

$$F_i = V_f \beta_i + \omega_f F_i^{old}, \quad \beta_i = \beta_i^{food} + \beta_i^{i,best} \tag{13}$$

Where the variables are the foraging speed ($V_f$), the inertia weight ($\omega_f \in [0\ 1]$), previous foraging motion ($F_i^{old}$), the food attraction and the effect of the best fitness of the $i$th krill so far ($\beta_i^{food}$, $\beta_i^{i,best}$) which are defined as:

$$\beta_i^{food} = C^{food}\widehat{K}_{i,food}\widehat{X}_{i,food}, \quad \beta_i^{i,food} = C^{food}\widehat{K}_{i,i\ best}\widehat{X}_{i,i\ best} \tag{14}$$

The food coefficient, denoted as $C^{food}$, is a variable that changes during the iteration process using a uniformly distributed random variable ($rand \in [0\ 1]$).

$$C^{food} = 2\left(rand + \frac{g}{g_{max}}\right) \tag{15}$$

### 4.2.3. Random Diffusion

The dispersion of krill individuals in their physical environment can be explained by the maximum speed of diffusion, combined with a randomly determined directional vector.

$$D_i = D^{max}\left(1 - \frac{g}{g_{max}}\right)\delta \tag{16}$$

where $D^{max}$ denotes the maximum diffusion speed and $\delta$ represents uniformly distributed random vector between $-1$ and $1$.

### 4.2.4. Updating Position

The three factors mentioned earlier prompt each krill individual to modify its position in alignment with the optimal direction. The adjustment of an individual's position during the time interval $t + \Delta t$ can be represented by the following expression:

$$X_i(t + \Delta t) = X_i(t)\Delta t \frac{dX_i}{dt} \tag{17}$$

The $\Delta t$ is crucial and its value entirely relies on the characteristics of the search space. It can be represented as:

$$\Delta t = C_t \sum_{i=1}^{NV}(UB_i - LB_i) \tag{18}$$

The equation is determined by the constant $C_t$, which is a number between 0 and 2. $NV$ denotes the overall count of control variables, whereas $UB_i$ and $LB_i$ represent the upper and lower boundaries of the jth variable, respectively.

### 4.2.5. Genetic Operators

To enhance the performance of the KH algorithm, the crossover and mutation strategies of the Genetic Algorithm are integrated. The crossover operation is formulated as follows:

$$X_{i,j} = \begin{cases} X_{r1,j} & if \quad rand < C_R \\ X_{i,j} & else \end{cases}, \quad j = 1, \dots, D \quad I = 1, \dots, NP \quad (19)$$

$$C_R = 0.05/\widehat{K}_{i,best} \quad (20)$$

where $D$ represents the dimension of the optimal problem, $X_{r1}$ ($r1 \neq i$) is randomly selected from the current population, $C_R$ denotes the probability of crossover. For the global best solution, $C_R$ is set to zero. The mutation is applied as follows:

$$X_{i,j} = \begin{cases} X_{best,j} + \mu(X_{r2,j} - X_{r3,j}) & if \quad rand < Mu \\ X_{i,j} & else \end{cases},$$
$$j = 1, \dots, D \quad i = 1, \dots, NP \quad (21)$$

$$Mu = 0.05/\widehat{K}_{i,best} \quad (22)$$

In this context, $X_{best}$ refers to the overall best position within the entire swarm, while $\mu$ represents the mutant factor that spans a range of values from 0 to 1. Additionally, factor, $X_{r2}$, and $X_{r3}$ (where $r2 \neq r3 \neq i$) are selected randomly from the present population. The mutant probability, denoted as $Mu$, is set to zero for the global best solution as well.

### 4.2.6. The Procedure of KHA

The KHA (Krill Herd Algorithm) can generally be defined by the following steps:

***Step 1.* Initialization:** Set the randomly generated initial population of krill individuals, define the search space boundaries ($X_{min}$ and $X_{max}$), and initialize the algorithm's parameters. Random values are assigned to each $D$-dimensional individual according to:

$$X_{j,i|g=0} = X_{j,min} + rand \times (X_{j,max} - X_{j,min}), \quad j = 1,..,D; \; i = 1,..,NP \quad (23)$$

where NP represents the population size.

***Step 2*. Fitness evaluation:** Assess the fitness of each krill individual in the population based on the objective function and save the global best solution.

***Step 3*. Motion calculation:**

    Movement induced by other krill individuals

    Foraging motion

    Random diffusion

***Step 4*. Crossover and Mutation:** Incorporate the crossover and mutation strategies from Genetic Algorithms to further improve the algorithm's performance.

***Step 5*.** Update the position of each krill individual and repeat the Step 2.

***Step 6*.** Check if the termination criteria have been met considering the termination criteria. If not, return to step 3.

## 4.3. Thermal Exchange Optimizer

In the Thermodynamics-Inspired Optimization (TEO) algorithm [63], a subset of agents is designated as cooling objects, while the remaining agents represent the environment. Interestingly, in TEO, this assignment is done contrariwise compared to traditional approaches. The algorithm follows the steps outlined below:

***Step 1*. Initialization:** In an m-dimensional search space, the initial temperature of all the objects is established.

$$T_i^0 = T_{min} + rand(T_{max} - T_{min}) \tag{24}$$

$T_i^0$ represents the initial solution vector of the $i$th object. $T_{min}$ and $T_{max}$ are the lower and upper bounds of the design variables, respectively, and $n$ denotes the total number of objects.

***Step 2*. Evaluation:** The objective function computes the cost value for each object.

***Step 3*. Saving:** In order to enhance the algorithm's performance without significantly increasing computational cost, a memory component is introduced to store historically best $T$ vectors along with their corresponding objective function values. This memory, referred to as the Thermal Memory ($TM$), is utilized in this step. The saved solution vectors in $TM$ are added to the population, while an equal number of the current worst objects are removed. Subsequently, the objects are sorted based on their objective function values in ascending order. This process helps incorporate valuable historical information into the population and maintain a diverse set of solutions.

*Step 4.* **Creating groups:** The agents in the population are divided into two equal groups. The pairs of agents are defined [63]. For example, $T_1$ serves as an environment object for $T_{2n+1}$, which acts as a cooling object, and vice versa. This pairing scheme ensures the interaction and exchange of heat between the environment and cooling objects in a structured manner.

*Step 5.* **Defining $\beta$:** In nature, when an object has a lower $\beta$ value, it tends to undergo only minor temperature exchanges. Drawing inspiration from this characteristic, a similar formulation is proposed in the algorithm. The value of $\beta$ for each object is evaluated using Eq. (25). In this equation, objects with lower cost values have lower $\beta$ values, indicating that they undergo smaller changes in position. This approach allows objects with better fitness (lower cost) to make gradual adjustments while exploring the search space.

$$\beta = \frac{Cost(object)}{Cost(worst\ object)} \tag{25}$$

*Step 6.* **Defining $t$:** The value of time, denoted as $t$, is associated with the iteration number in the formulation. The calculation of $t$ for each agent is determined using Eq. (226), which is given as:

$$t = \frac{iter.}{Max\_iter} \tag{26}$$

*Step 7.* **Escaping from local optima ($i$):** Metaheuristic algorithms should possess the capability to escape from traps encountered when agents approach local optima. Step 7 and Step 9 are employed specifically for this purpose. In these steps, the environmental temperature is adjusted using Eq. (27):

$$T_i^{env.} = (1 - (c_1 + c_2 \times (1-t)) \times rand) \times T'^{env.}_i \tag{27}$$

The previous temperature of the object, denoted as $T_i^{env.}$, is adjusted to a new temperature, $T'^{env.}_i$. The parameter $(1-t)$ is employed to reduce randomness as the iterations advance. As $t$ increases towards the end of the process, randomness decreases linearly, promoting exploitation. $c_2$ controls the factor $(1-t)$. For example, when decreasing is not required, $c_2$ can be set to zero. $c_1$ controls the magnitude of random steps. Additionally, when a decreasing process is not employed ($c_2 = 0$, as mentioned earlier), $c_1$ introduces randomness.

When $C = 0$ (when $c_1 = c_2 = 0$), none of the mechanisms mentioned above are applied, and the previous temperature is multiplied by 1. In this chapter, $c_1$ and $c_2$ are selected from the set {0 or 1}.

***Step 8*. Updating the agents:** Based on the previous steps, the new temperature of each object is updated using the following equation:

$$T_i^{env.} = T_i^{env.} + \left(T_i^{old} - T_i^{env.}\right)\exp(-\beta t) \tag{28}$$

***Step 9*. Escaping from local optima (ii):** The parameter Pro, which takes a value within the range of (0, 1), is introduced to determine whether a component of each cooling object should be changed or not. For each agent, Pro is compared with $Rand(i)$, where $i$ ranges from 1 to $n$ and $Ran(i)$ is a random number uniformly distributed within the interval (0, 1). If $Rand(i)$ is less than Pro, indicating a successful comparison, one dimension of the $i$th agent is randomly selected, and its value is regenerated using the following process:

$$T_{i,j} = T_{j,min} + rand\left(T_{j,max} - T_{j,min}\right) \tag{29}$$

The equation includes $T_{i,j}$, which represents the jth variable of the $i$th agent. $T_{j,min}$ and $T_{j,max}$ indicate the lower and upper limits, respectively, of the jth variable. To maintain the integrity of the agents' structures, only one dimension is modified in this process. By employing this mechanism, the agents are able to explore the entire search space, facilitating improved diversity and enhancing the likelihood of discovering optimal solutions.

***Step 10*. Checking terminating conditions:** The optimization process is concluded after a predetermined number of iterations. If the termination criterion is not met, the algorithm returns to step 2 to initiate a new round of iterations. However, if the termination criterion is satisfied, the process is stopped, and the best solution found throughout the optimization process is reported as the final result.

## 5. Results and Discussion on Case Studies: Truss Bridges

Truss bridges are a specific type of trusses that are suitable for shape and size optimization. These trusses are utilized in various bridge components, including the deck [64] and arch [65-68]. Topology or shape optimization plays a significant role in enhancing the structural form of truss bridges. By optimizing the shape, material usage can be significantly reduced while improving overall performance. Therefore, truss bridges are classified as the third category among the truss benchmark examples, which include the Michell arch, forth bridge model, and the 37-bar truss bridge. The design data pertaining to the optimization problems for these examples can be found in Table 2.

**Table 2.** Design data for truss bridges.

| | Michell Arch | Forth bridge model | 37-bar truss bridge |
|---|---|---|---|
| Variables (Size/shape) | $A_1 = A_8; A_2 = A_7;$ $A_3 = A_6; A_4 = A_5;$ $A_9 = A_{13}; A_{10} = A_{12}; A_{11};$ $x_3 = -x_7; y_4 = y_6; y_5;$ | $A_1, A_2 \ldots, A_{16}$ $y_i$ | $A_1, A_2 \ldots, A_{14}$ $y_{upper\ chord}$ |
| Constraint (Stress, disp., freq., shape) | $\|(\sigma)_{i=1\sim13}\| \leq 240$ MPa; $\Delta_{y(1)} \leq 3.8$ mm $0 \leq x_3 \leq 1$ m; $0 \leq y_4, y_5 \leq 1$ m; | $\|(\sigma)_i\| \leq 25\ kN/cm^2$; $-1.4 \leq \Delta_{y(i)} \leq 1.4$ m; | $f_1=20, f_2=40, f_3=60$ (Hz) $-1 \leq \Delta_{y(top\ chord)} \leq 2.5$ m; |
| Cross-sections | $A_i = \{1.01, 1.02, \ldots, 5\}cm^2$ $i = 1, 2, \ldots, 13$ | $0.5 \leq A_i \leq 100 cm^2$ $i = 1, 2, \ldots, 16$ | $A_{top\ chord} = 4 \times 10^{-3}\ m^2$ $10^{-4} \leq A_{others} \leq 3.5 \times 10^{-4}\ m^2$ |
| Load case ($F_y$) | P=-200 kN (Node 1) | P=20 ton | $Mass_i=10$ kg $i$ = nodes on lower chord |
| Young's modulus | 210 GPa | $2.1 \times 10^8\ kN/m^2$ | $2.1 \times 10^{11}\ N/m^2$ |
| Material's density | 7800 kg/m$^3$ | 7800 $kg/m^3$ | 7800 $kg/m^3$ |

### 3.3.1 The Michell Arch

The Michell arch, as shown in Figure 1, is recognized as the initial instance of its kind. This optimization problem has an available analytical solution that takes into account equal allowable stresses in both tension and compression (Wang et al., 2002):

$$W = \frac{12}{\sigma^+} LP\rho \tan\frac{\pi}{12}$$

In the equation, the parameter L represents half of the span length, specifically set to 1m, while $\sigma^+$ denotes the allowable tension stress. Table 3 presents the optimal outcome obtained from the memory-assisted methods, along with a comparative analysis against other approaches. Considering the results of the memory-less version of algorithms (21.91, 22.2, and 23.88 kg), it is apparent that memory-assisted versions (21.21, 20.96, and 22.03 kg) performed better and found minimal weight while satisfying all the constraints. Figure 1 visually displays the optimal arrangement of the structure's components. Additionally, Fig. 2 showcases the best convergence of the memory-assisted methods, among 20 individual runs for the Michell arch. Considering results, the memory-assisted algorithms are considerably reliable and efficient with fast convergence. The average improvement of 5.5% (max 7.7%) is shown in the achieved weights.

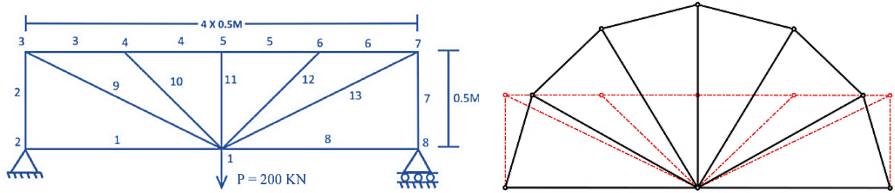

**Fig. 1** The Michell arch

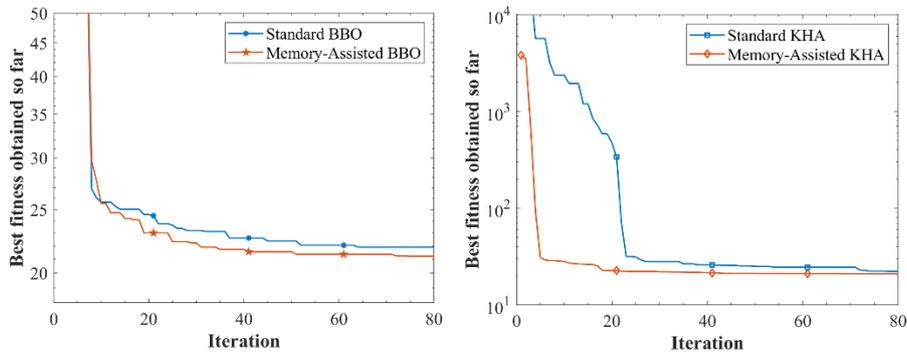

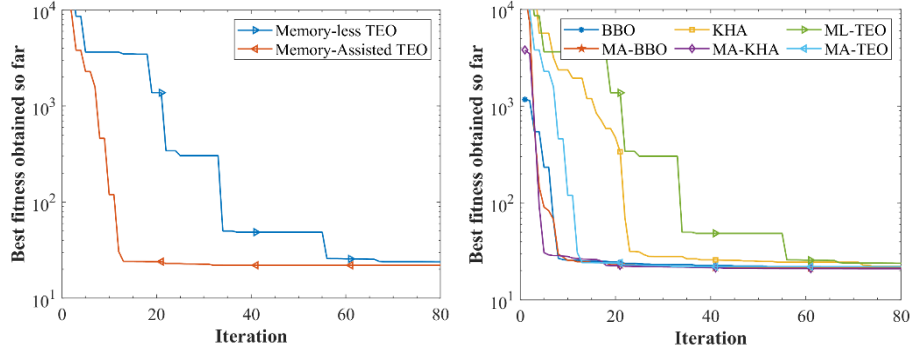

**Fig. 2**. Results of the Michell arch

**Table 3.** Results comparison of the standard and memory-based methods for Michell arch problem.

| Variables | BBO | | KHA | | TEO | |
|---|---|---|---|---|---|---|
| | Standard | Memory | Standard | Memory | Standard | Memory |
| Sizing Variables (m$^2$) | | | | | | |
| $A_1$ | 23.1535 | 14.1476 | 1.03960 | 14.7664 | 66.9590 | 62.6198 |
| $A_2$ | 339.7858 | 335.4370 | 383.5986 | 333.5205 | 373.4588 | 366.8499 |
| $A_3$ | 361.0238 | 339.1281 | 331.4342 | 330.6981 | 368.7377 | 343.6751 |
| $A_4$ | 336.0014 | 339.5899 | 360.4571 | 327.9014 | 356.3795 | 342.0028 |
| $A_9$ | 153.0759 | 120.4642 | 144.4969 | 106.923 | 168.5932 | 47.2726 |
| $A_{10}$ | 118.2371 | 144.2717 | 165.5149 | 161.0867 | 186.3108 | 257.6826 |
| $A_{11}$ | 129.7560 | 103.8559 | 83.3975 | 76.1589 | 147.8932 | 24.8989 |
| Layout Variables (m) | | | | | | |
| $Y_5$ | 1.0000 | 1.0000 | 0.9907 | 0.9996 | 0.9513 | 0.9534 |
| $Y_6$ | 0.8680 | 0.8795 | 0.8916 | 0.8951 | 0.8345 | 0.8863 |
| $X_3$ | 0.8862 | 0.8665 | 0.8820 | 0.8648 | 0.8548 | 0.8141 |
| Statistical results (kg) | | | | | | |
| Best | 21.91 | 21.21 | 22.2 | 20.96 | 23.88 | 22.03 |
| Mean | 22.23 | 21.66 | 23.21 | 22.16 | 1971.54 | 1949.11 |
| Worst | 22.58 | 22.28 | 25.01 | 24.67 | 16533.85 | 33298.82 |
| Std | 0.18 | 0.27 | 0.7 | 1.08 | 4208.62 | 7403.29 |
| NFEs | 3200 | 3750 | 3900 | 3900 | 4000 | 4000 |
| Runs | 20 | 20 | 20 | 20 | 20 | 20 |

### 3.3.2 The Forth bridge model

Gil and Andreu (2001) initially analyzed the Forth bridge model to determine the optimal configuration and sizing variables. Each span in the bridge has a length of 16m and a height of 1m (adjusted to 3m to control shape variables). The structure can be represented according to Figure 3, which depicts half of the infinite symmetric span. The circled numbers in Figure 3(b) indicate 16 groups of sizing variables. By employing a memory-assisted optimization process, the best structural weight achieved was found to be significantly lower than that obtained using memory-less algorithms. Specifically, the weights were 11978.62 kg, 10349.48 kg, and 15270.36 kg, compared to 13585.99 kg, 11775.9 kg, and 21629.57 kg for memory-less algorithms. Table 4 provides the relevant coordinates and cross-sections for the optimized structure.

It is crucial to emphasize the importance of increasing the overall moment of inertia at the support positions of the bridge to withstand the high internal moments experienced during its operational lifespan. Therefore, a meticulous analysis and optimization of the support positions are essential to ensure the structural integrity and capacity to withstand anticipated loads and stresses. The optimal shape, as depicted in Fig. 4(a), satisfies this requirement and bears resemblance to the well-known "Forth Bridge." The best convergence histories of 20 individual runs are shown in Fig. 4(b). The memory-assisted algorithms have demonstrated significant performance improvements compared to the memory-less algorithms. On average, the optimization results have improved by 17.7%, with a maximum improvement of 29.4%. Additionally, the convergence behavior has been found to be superior in the memory-assisted algorithms.

**Table 4.** Results comparison of the standard and memory-based methods for the Forth bridge model

| Variables | BBO | | KHA | | TEO | |
|---|---|---|---|---|---|---|
| | Standard | Memory | Standard | Memory | Standard | Memory |
| | Sizing Variables (cm$^2$) | | | | | |
| 1 | 16.4456 | 25.1129 | 14.836 | 10.538 | 37.3404 | 18.7656 |
| 2 | 94.254 | 38.4602 | 47.9999 | 41.0708 | 42.049 | 38.6652 |
| 3 | 15.8908 | 7.4326 | 9.0737 | 9.9726 | 46.0009 | 14.7888 |
| 4 | 39.2538 | 46.8004 | 55.3138 | 57.9283 | 28.0406 | 44.4536 |
| 5 | 56.6392 | 37.434 | 34.6687 | 26.8824 | 31.8419 | 36.5998 |
| 6 | 22.0073 | 26.2891 | 24.3291 | 16.7202 | 24.5844 | 24.6589 |
| 7 | 0.5 | 34.3952 | 33.3137 | 40.5562 | 11.5118 | 17.4166 |
| 8 | 14.8243 | 0.5 | 8.7103 | 0.5289 | 17.1568 | 2.7853 |
| 9 | 25.5838 | 30.9563 | 28.1914 | 31.3519 | 20.5659 | 24.8771 |
| 10 | 40.2238 | 0.5 | 1.6936 | 0.6179 | 11.8985 | 10.1461 |
| 11 | 7.0234 | 31.5138 | 24.2553 | 32.994 | 10.4383 | 18.0454 |
| 12 | 14.209 | 13.8402 | 16.0194 | 12.6324 | 13.7095 | 15.9911 |
| 13 | 23.8645 | 27.8096 | 33.2668 | 34.5302 | 20.149 | 34.7149 |
| 14 | 9.0283 | 4.9822 | 10.9916 | 3.5478 | 22.1609 | 12.8686 |
| 15 | 47.5124 | 12.1978 | 13.6822 | 12.4636 | 25.4358 | 13.1036 |
| 16 | 4.9522 | 13.9933 | 9.6314 | 0.5187 | 35.8006 | 11.8662 |

|   | Layout Variables (cm) | | | | | |
|---|---|---|---|---|---|---|
| 1 | -57.2705 | 73.6468 | 0.1716 | 36.1304 | -12.6051 | 30.6401 |
| 2 | 37.5484 | 66.0000 | 65.6483 | 101.3377 | 3.4807 | 52.8938 |
| 3 | -102.441 | -13.2346 | -54.181 | -29.985 | -1.7244 | -22.3989 |
| 4 | 90.7979 | 140.000 | 128.5676 | 139.9841 | 42.6357 | 109.1288 |
| 5 | -109.620 | -54.6006 | -72.3198 | -98.6304 | -41.0543 | -61.3265 |
| 6 | 140.000 | 89.2177 | 58.9762 | 90.7673 | 24.2037 | 42.0780 |
| 7 | -50.5744 | -140.00 | -73.9872 | -140.000 | -16.0522 | -0.1548 |
| 8 | 33.6759 | 21.9731 | 27.5776 | 32.5376 | -0.0001 | 12.5168 |
| 9 | -12.7194 | -81.1674 | -41.4697 | -71.1337 | -24.7971 | -8.1538 |
| 10 | 44.6205 | 5.2223 | 37.6646 | 20.2854 | 25.8910 | 25.9354 |
|   | Statistical results (kg) | | | | | |
| Best | 13585.99 | 11978.62 | 11775.9 | 10349.48 | 21629.57 | 15270.36 |
| Mean | 15021.67 | 14695.9 | 13867.89 | 12082.04 | 24069.96 | 18888.7 |
| Worst | 17233.06 | 16705.82 | 16946.59 | 13544.55 | 27361.33 | 21795.11 |
| Std | 1026.5 | 1167.75 | 1286.29 | 786.42 | 1562.61 | 1646.89 |
| NFEs | 3800 | 3200 | 4000 | 4000 | 1950 | 2300 |
| Runs | 20 | 20 | 20 | 20 | 20 | 20 |

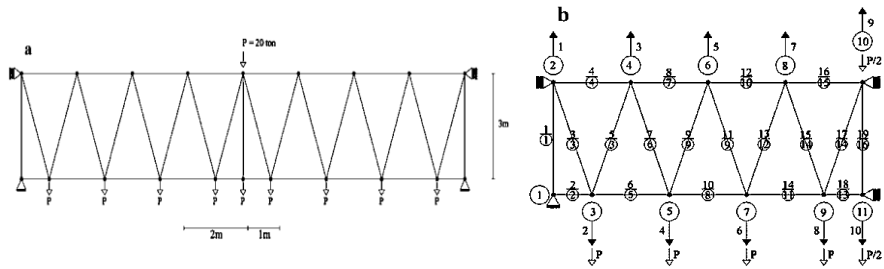

**Fig. 3** The Forth bridge model (a) Problem diagram. (b) Analytical model.

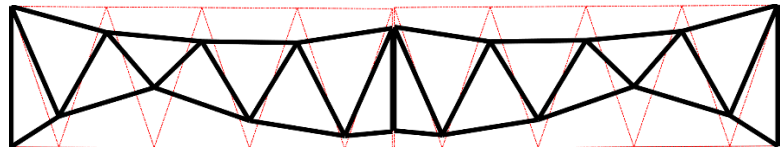

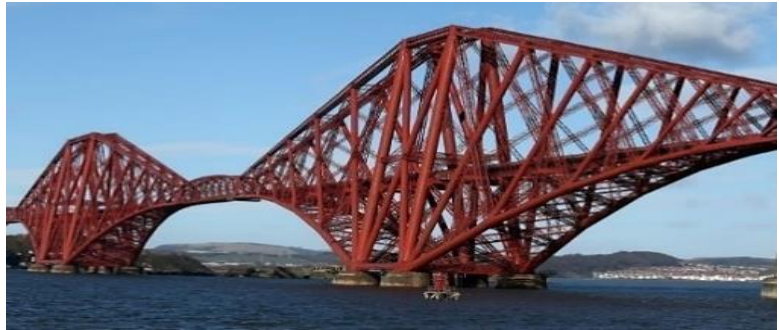

**a)** Optimum and real shape

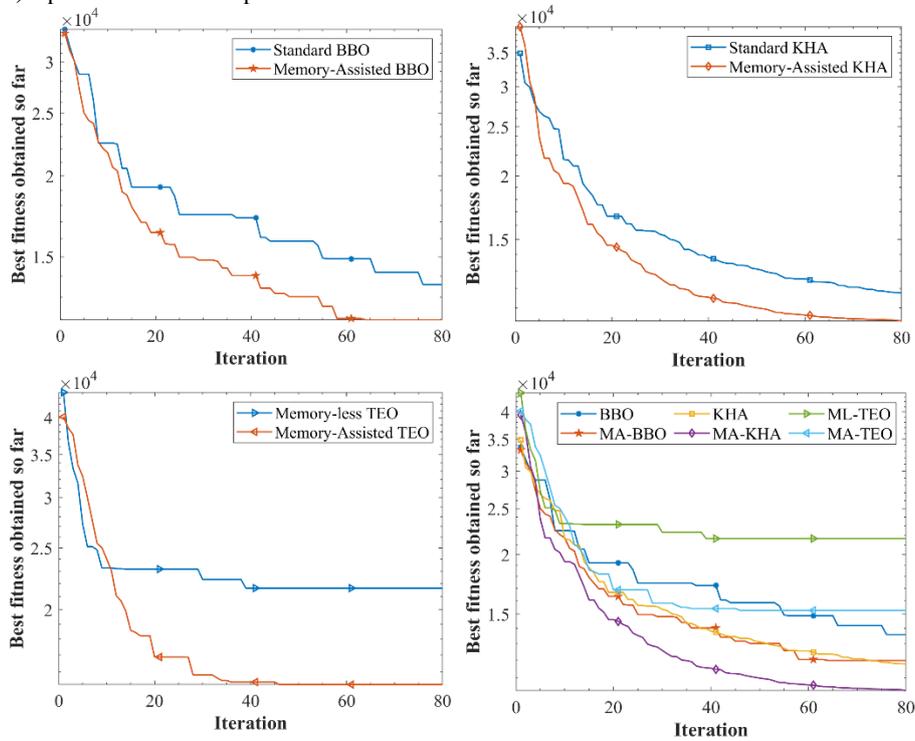

**b)** Convergence history of methods
**Fig. 4**. Results of the Forth bridge model

### *3.3.3 The 37-bar truss bridge*

The simply supported Pratt-Type 37-bar truss (depicted in Figure 5(a)) has been investigated by Lingyun et al. (2005) using the NHGA algorithm and by Wang et al. (2004) employing the method of evolutionary node shift. This truss presents an optimization problem with 14 sizing variables, five shape variables, and three frequency constraints. Table 5 provides a comparison between the optimal results obtained using the memory-assisted methods and their standard versions. The memory-assisted algorithms have yielded diverse levels of performance in optimizing the design, with the best results achieved being 357.81 kg, 356.05 kg, and 367.19 kg, in contrast to the values of 361.1 kg, 359.04 kg, and 378.21 kg obtained by memory-less algorithms. On average, the utilization of memory has improved performance by 1.5%, with a maximum improvement of 2.9%. This improvement can be attributed to the incorporation of elite population memory, allowing the enhanced algorithms to explore the search space more effectively and converge to superior solutions at a faster pace. Figure 5(b) showcases the final optimal shape of the truss, while Figure 5(c) illustrates the weight convergence of the 37-bar truss for the MAMVO method, which exhibits better performance when compared to MVO.

**Table 5.** Results comparison of the standard and memory-based methods for the 37-bar truss problem.

| Variables | BBO | | KHA | | TEO | |
|---|---|---|---|---|---|---|
| | Standard | Memory | Standard | Memory | Standard | Memory |
| | | | Sizing Variables ($m^2$) | | | |
| $Y_3, Y_{19}$ | 1.1204 | 1.0335 | 1.0135 | 1.0719 | 1.1572 | 1.0000 |
| $Y_5, Y_{17}$ | 1.6069 | 1.3060 | 1.4347 | 1.4280 | 1.9219 | 1.3569 |
| $Y_7, Y_{15}$ | 1.7317 | 1.7173 | 1.8022 | 1.6202 | 1.9554 | 1.5958 |
| $Y_9, Y_{13}$ | 1.8849 | 1.9628 | 1.9338 | 1.7794 | 2.0202 | 1.6577 |
| $Y_{11}$ | 2.0734 | 1.9995 | 2.1250 | 1.8409 | 1.9787 | 1.7288 |
| | | | Layout Variables (cm) | | | |
| $A_1, A_{27}$ | 2.1053 | 1.8660 | 2.3616 | 2.5027 | 2.4087 | 2.0580 |
| $A_2, A_{26}$ | 1.4703 | 1.1156 | 1.0303 | 1.0003 | 1.8259 | 1.3164 |
| $A_3, A_{24}$ | 1.0000 | 1.0000 | 1.1233 | 1.1655 | 1.3396 | 1.8695 |
| $A_4, A_{25}$ | 2.1949 | 2.5251 | 2.7226 | 2.1397 | 2.1313 | 2.7549 |
| $A_5, A_{23}$ | 1.0000 | 1.0161 | 1.0889 | 1.0234 | 1.9914 | 1.5808 |
| $A_6, A_{21}$ | 1.0000 | 1.2085 | 1.2246 | 1.0692 | 2.7621 | 1.4882 |
| $A_7, A_{22}$ | 2.3970 | 2.1943 | 1.3502 | 1.8263 | 1.8349 | 3.4876 |
| $A_8, A_{20}$ | 1.2092 | 1.0000 | 1.0365 | 1.005 | 1.9548 | 1.3113 |
| $A_9, A_{18}$ | 1.1588 | 1.0000 | 1.3631 | 1.0055 | 1.8608 | 1.6453 |
| $A_{10}, A_{17}$ | 2.4954 | 2.7481 | 2.3905 | 2.1666 | 1.4984 | 2.1387 |
| $A_{11}, A_{19}$ | 1.5297 | 1.0248 | 1.4385 | 1.1465 | 1.5211 | 1.8052 |
| $A_{12}, A_{15}$ | 1.4765 | 1.15 | 1.1589 | 1.0457 | 1.2811 | 1.3022 |
| $A_{13}, A_{16}$ | 2.2398 | 2.0232 | 2.1901 | 3.4073 | 2.6857 | 2.7191 |
| $A_{14}$ | 1.4364 | 1.0000 | 1.0122 | 1.0000 | 2.2287 | 2.0468 |
| | | | Statistical results (kg) | | | |
| Best | 361.1 | 357.81 | 359.04 | 356.05 | 378.21 | 367.19 |

| Mean | 363.24 | 361.64 | 362.79 | 359.38 | 443.14 | 387.61 |
| Worst | 366.18 | 365.53 | 365.97 | 363 | 608.5 | 652.68 |
| Std | 1.14 | 2 | 1.98 | 1.87 | 81.64 | 62.67 |
| NFEs | 3400 | 3950 | 3750 | 3600 | 3500 | 2600 |
| Runs | 20 | 20 | 20 | 20 | 20 | 20 |

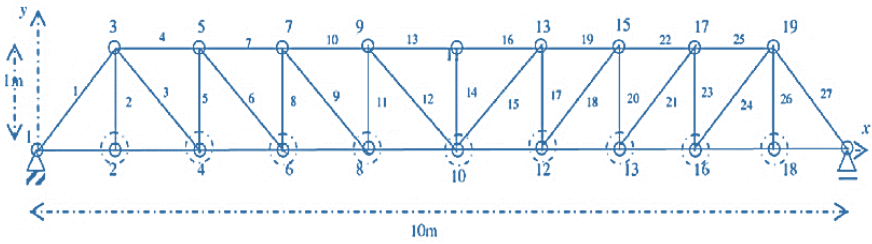

**a)** The 37-bar truss

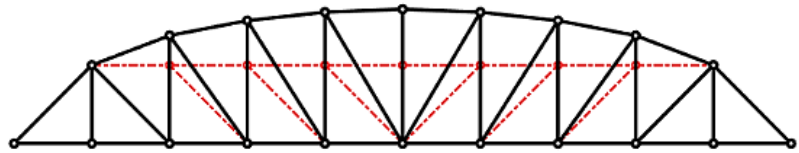

**b)** The optimum 37-bar truss

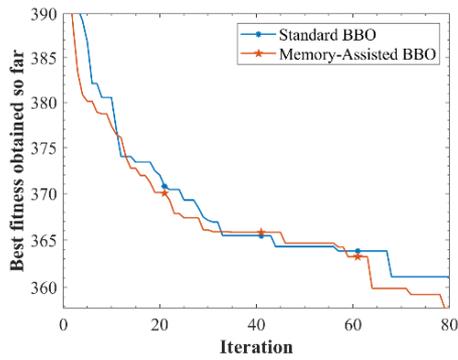 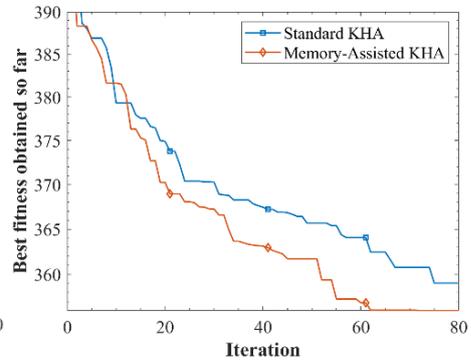

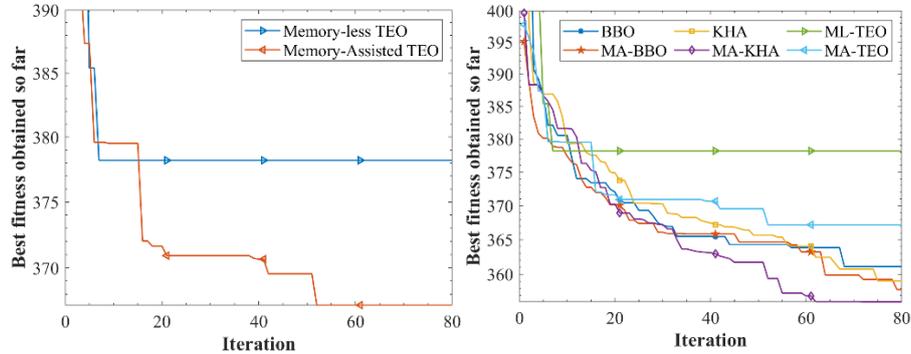

**b)** Convergence history of methods
**Fig. 5**. Results of the 37-bar truss

## 4. Conclusions

In this chapter, the effectiveness of utilizing memory-based techniques in enhancing the performance of various metaheuristic algorithms has been investigated. By adding a memory component to the basic optimization algorithm, several best solutions from previous iterations are saved to be exchanged with several worst solutions in the next iterations. This strategy is called a multi-elite strategy and has been applied to a number of well-known optimization algorithms such as the Biogeography-Based Optimization (BBO), Krill Herd Algorithm (KHA), and Thermal Exchange Optimization (TEO). The proposed memory implementation allowed the algorithm to avoid the worst solutions by exchanging them (20% of the population size) with the global best ones from the memory. In this chapter, the proposed memory-assisted versions of these algorithms have been implemented to optimize benchmark engineering examples, such as size and shape optimization of truss bridges. The results show that the memory-assisted optimization algorithms outperform the memory-less ones in terms of convergence speed, solution quality, and reliability. The case study examples of truss bridges demonstrated that the proposed method can effectively optimize problems with both continuous and discrete variables. Therefore, utilizing memory-based techniques can be a promising approach to improve the performance of various optimization algorithms and to solve complex engineering problems.